\pdfoutput=1

\documentclass[11pt]{article}

\usepackage[preprint]{acl}

\usepackage{makecell}
\usepackage{times}
\usepackage{latexsym}

\usepackage[T1]{fontenc}

\usepackage[utf8]{inputenc}

\usepackage{microtype}

\usepackage{inconsolata}

\usepackage{graphicx}
\usepackage{amsmath,amsfonts}
\usepackage{algorithmic}
\usepackage{algorithm}
\usepackage{array}
\usepackage[caption=false,font=normalsize,labelfont=sf,textfont=sf]{subfig}
\usepackage{textcomp}
\usepackage{stfloats}
\usepackage{url}
\usepackage{verbatim}
\usepackage{caption}
\usepackage{graphicx}
\usepackage{times}
\usepackage{soul}
\usepackage{url}
\usepackage[utf8]{inputenc}
\usepackage[hypcap=false]{caption}
\usepackage{graphicx}
\usepackage{amsmath}
\usepackage{bbding}
\usepackage{pifont}
\usepackage{amsthm}
\usepackage{float}
\usepackage{booktabs}
\usepackage{algorithm}
\usepackage{algorithmic}
\usepackage{multirow}
\usepackage[table]{xcolor}
\usepackage{amssymb}
\definecolor{lightblue}{RGB}{236, 244, 255}
\definecolor{mylightgray}{RGB}{240, 240, 240}

\title{\textbf{\textit{MoIIE}}: Mixture of Intra- and Inter-Modality Experts for Large Vision Language Models}

\author{
    Dianyi Wang$^{1,2}$\enskip\enskip 
    Siyuan Wang$^{3}$\enskip\enskip
    Zejun Li$^{1}$\enskip\enskip
    Yikun Wang$^{1,2}$\enskip\enskip\\
    {\bfseries Yitong Li$^{4}$}\enskip\enskip
    {\bfseries Duyu Tang$^{4}$}\enskip\enskip
    {\bfseries Xiaoyu Shen$^{5}$}\enskip\enskip
    {\bfseries Xuanjing Huang$^{1}$}\enskip\enskip
    {\bfseries Zhongyu Wei$^{1,2}$}\\
    \textsuperscript{1}Fudan University \enskip 
    \textsuperscript{2}Shanghai Innovation Institut \enskip \\
    \textsuperscript{3}University of Southern California \enskip 
    \textsuperscript{4}Huawei Technologies Co., Ltd \enskip \\
    \textsuperscript{5}Ningbo Key Laboratory of Spatial Intelligence and Digital Derivative, Institute of Digital Twin, EIT \enskip \\
    \texttt{dywang24@m.fudan.edu.cn, sw\_641@usc.edu}
}

\begin{document}
\maketitle
\begin{abstract}
Large Vision-Language Models (LVLMs) have demonstrated remarkable performance across multi-modal tasks by scaling model size and training data. However, these dense LVLMs incur significant computational costs and motivate the exploration of sparse Mixture of Experts (MoE) architectures. While MoE improve parameter efficiency, effectively applying MoE to simultaneously model modality-specific features and cross-modal associations in LVLMs remains challenging. In this work, we propose to incorporate Mixture of Intra- and Inter-Modality Experts (MoIIE) to LVLMs. For each token, expert routing is guided by its modality, directing tokens to their respective intra-modality experts as well as a shared pool of inter-modality experts, enabling the model to jointly learn rich intra-modal features and cross-modal interactions. We further introduce an effective and straightforward two-stage training strategy, which facilitates the direct activation of both MoE and multi-modal capabilities. Extensive experiments across different data scales, visual encoder and LLM backbone demonstrate the effectiveness, efficiency and generality of our approach. Notably, our MoIIE models with 5.5B and 11.3B activated parameters match or even surpass the performance of existing advanced open-source MoE-LLMs based multi-modal models that involve more activated parameters.
\end{abstract}

\section{Introduction}
\label{sec:intro}

Large Vision-Language Models (LVLMs)~\cite{bai2023qwen,dai2023instructblip,liu2023improved,minigpt-4} have gained significant attention for their ability to process information across both visual and linguistic modalities~\cite{cui2023surveymultimodallargelanguage,202411.0685}. By integrating visual encoders~\cite{radford2021clip,zhai2023siglip} with Large Language Models (LLMs) through connection module~\cite{ lin2024preservecompressindepthstudy}, LVLMs align high-dimensional visual features with the linguistic knowledge and reasoning capabilities of LLMs~\cite{bai2023qwenlan,vicuna2023,bi2024deepseek}, demonstrating effectiveness across diverse cross-modal tasks~\cite{liu2024mmbenchmultimodalmodelallaround,fu2024mmecomprehensiveevaluationbenchmark}. 

As with unimodal LLMs, scaling up model size has been shown to improve performance in multi-modal settings~\cite{liu2024llava,chen2024fargpt4vclosinggap} but also significantly increases computational costs, especially when using dense Transformer~\cite{vaswani2023attentionneed} architectures.
To maintain efficiency while scaling parameters, recent studies introduce Mixture of Experts (MoE)~\cite{lepikhin2020gshardscalinggiantmodels} into LLMs, replacing dense feed-forward network (FFN) layers with sparsely activated expert layers. This approach adaptively selects only a small subset of experts for each input based on token-level routing decisions, thus reducing computational overhead while enhancing model capacity. 

\begin{figure*}[t]
    \centering \includegraphics[width=0.8\linewidth]{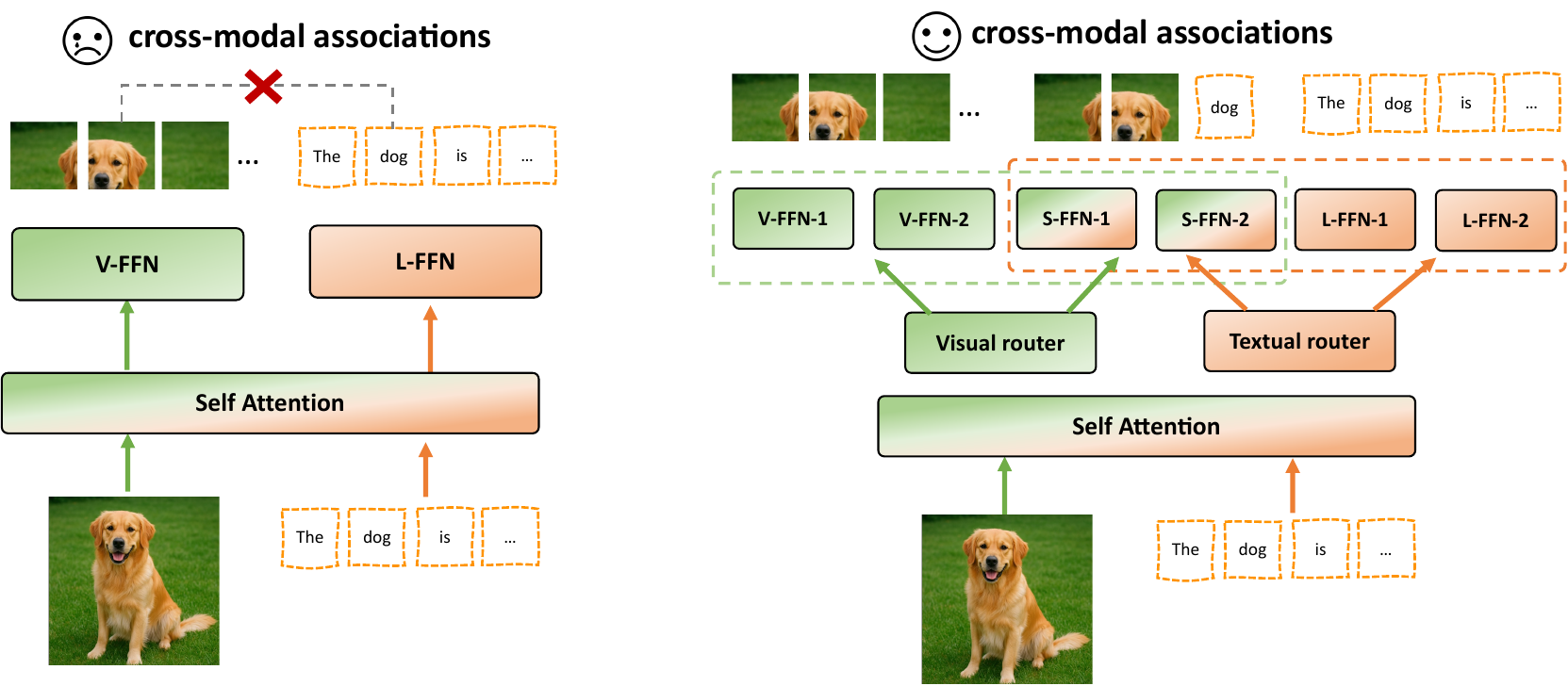}
    \caption{\textbf{Modality-specific MoE v.s. Our MoIIE.} \textbf{Left}: The modality-specific MoE module routes text and image tokens exclusively to their respective specialized expert groups, limiting cross-modal associations such as the alignment between ``dog'' token and its corresponding image region.
    \textbf{Right}: Our MoIIE introduces intra-modality and inter-modality expert groups. Intra-modality experts (Expert V for image tokens, Expert L for text tokens) process modality-specific features while inter-modality experts (Expert S) process tokens from both modalities to model cross-modal interactions. }
    \label{fig:intro}
\end{figure*}

For multi-modal MoE implementation, a common approach is to directly extend vanilla MoE designs from LLMs to LVLMs by routing tokens from all modalities to a shared pool of experts~\cite{lin2024moellavamixtureexpertslarge,han2025fusemoemixtureofexpertstransformersfleximodal}. However, this overlooks the fundamental differences in information density and feature distribution between text and image tokens~\cite{liang2023foundationstrendsmultimodalmachine}. An alternative approach incorporates modality-specific experts, where text and image tokens are routed to their respective specialized expert groups. While this design enables more specialized feature learning for each modality~\cite{lin2024momaefficientearlyfusionpretraining,wang2024cogvlmvisualexpertpretrained,shen2023scalingvisionlanguagemodelssparse}, these experts primarily focus on intra-modal knowledge and neglect cross-modal associations, such as the alignment between noun tokens in text and corresponding entity regions in images~\cite{xiao2024seeingimageprioritizingvisual}, as illustrated in Figure~\ref{fig:intro} (a).

To this end, we proposes the Mixture of Intra- and Inter-Modality Experts (MoIIE), to simultaneously capture both modality-specific features and cross-modal associations in LVLMs. As illustrated in Figure~\ref{fig:intro} (b), MoIIE comprises three distinct groups of experts: two intra-modality groups that specialize in language and vision, independently processing text and image tokens; and an inter-modality expert group shared by both modalities that focuses on cross-modal interactions between text and image tokens.
Correspondingly, MoIIE learns two dedicated routers respectively for image and text tokens, with each token dynamically routed to the most relevant experts from the corresponding intra-modality group and the inter-modality group. As not all tokens are strongly associated across modalities, such as function words in text or visual regions without descriptive content, the routing mechanism allows for flexible combinations. Some tokens may activate only intra-modality experts or only inter-modality experts, while others may access both  expert groups simultaneously. This dynamic routing strategy is consistently applied to both image and text tokens. Additionally, for multi-modal MoE training, most existing approaches adopt a sparse upcycling strategy~\cite{komatsuzaki2023sparseupcyclingtrainingmixtureofexperts} that transforms dense LVLMs into sparse models. This strategy generally follows a three-stage training pipeline: multi-modal pre-training of the dense model, fine-tuning the dense model on diverse multitasking datasets with all parameters unfrozen, and model sparsification with only expert modules trained~\cite{lin2024moellavamixtureexpertslarge,chen2024llavamolesparsemixturelora}.
However, this pipeline is relatively cumbersome and the limited training during the final stage potentially undermines the model’s generalization capabilities. In contrast, we propose that multi-modal fine-tuning and sparsification can be jointly optimized with all parameters updated in a more straightforward and effective two-stage training strategy. In the first stage, visual inputs are aligned with the LLM backbone through multi-modal pre-training datasets. In the second stage, MoIIE is integrated and the entire model is optimized using fine-tuning datasets, to simultaneously learn (1) versatile multi-modal understanding and instruction-following capabilities, while enabling (2) expert learning of modality-specific knowledge with cross-modal association within the MoIIE module. Our experimental results demonstrate that this simplified strategy is easy to implement and also effectively enhances performance across diverse downstream tasks.


In summary, our contributions are threefold:
\begin{itemize}
    \item We propose a robust sparse LVLM framework equipped with the novel MoIIE module, effectively modeling both modality-specific features and cross-modal associations.
    \item We introduce an effective and straightforward two-stage training strategy that simultaneously optimizes multi-modal fine-tuning and MoE modules.
    \item We conduct extensive experiments demonstrating that our MoIIE-integrated model through the two-stage training strategy achieves superior scaling efficiency compared to existing dense, modality-expert-based, and original MoE-based LVLMs.
\end{itemize}


\begin{figure*}[t]
    \centering
    \includegraphics[width=0.8\linewidth]{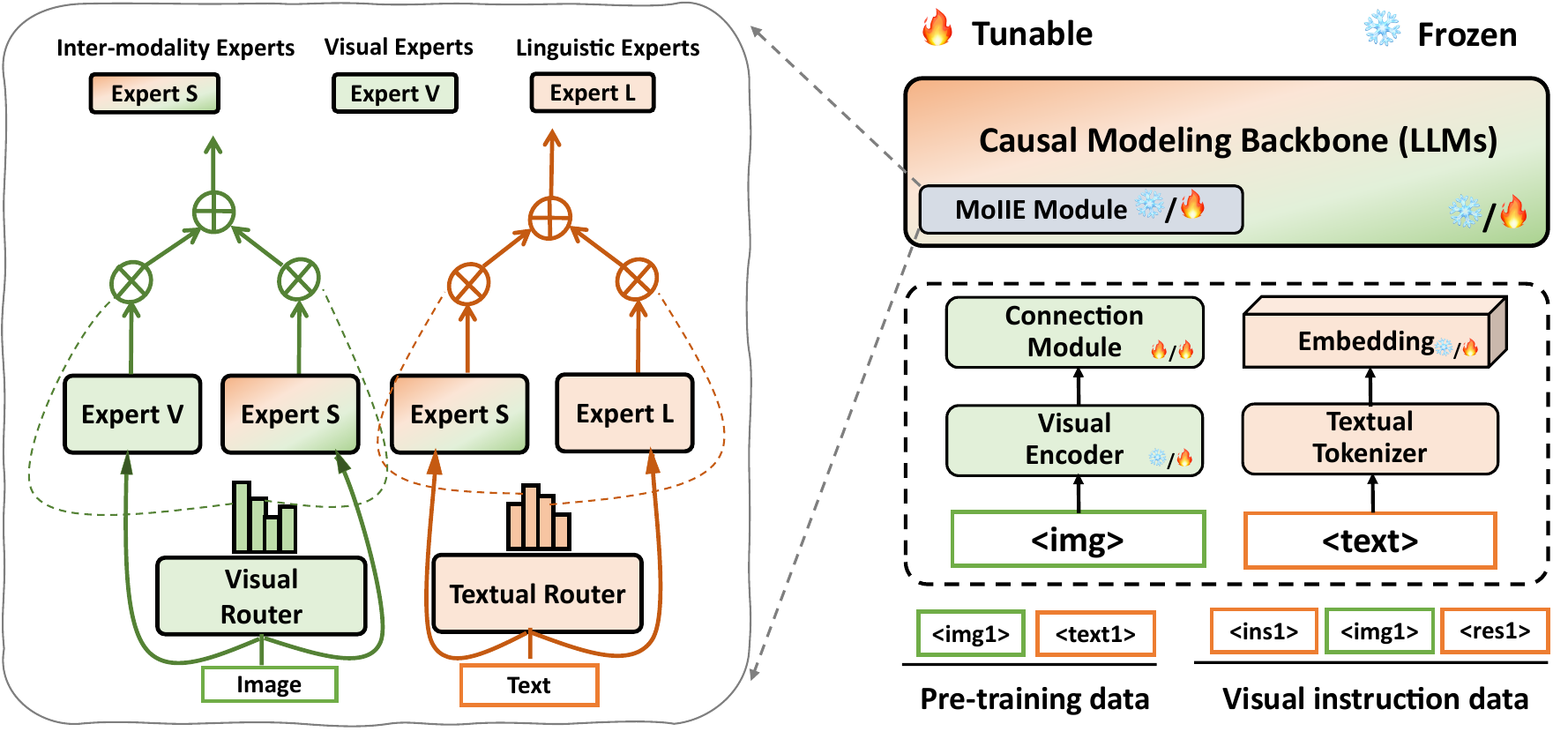} 
    \caption{\textbf{Method overview}. \textbf{Left}: The MoIIE architecture, consisting of intra-modality experts (Expert V for image tokens and Expert L for text tokens) and inter-modality experts (Expert S) that process tokens from both modalities. \textbf{Right}: The two-stage training strategy. For each module, the tunable or frozen icon before the slash indicates the configuration during the first stage, while the icon after the slash represents the second-stage setup.}
    \label{fig:MoIIE}
\end{figure*}

\section{Methodology}

\subsection{Overview}
We introduce MoIIE, a multi-modal MoE variant integrated into LVLMs to capture both modality-specific features and cross-modal associations, with the comprehensive architecture outlined in Figure~\ref{fig:MoIIE}. 
It incorporates a pretrained visual encoder coupled with a connection module that transforms visual inputs into sequential representations matching the dimension of LLM token embeddings. A detailed explanation of the MoIIE architecture is provided in Section \ref{section:Architecture}, followed by our proposed simple two-stage training paradigm in Section \ref{section:training}.
A detailed comparison between the original MoE and our MoIIE is shown in Appendix, highlighting the key architectural innovations improving multi-modal learning.

\subsection{MoIIE Architecture}
\label{section:Architecture}

\subsubsection{Multi-modal Input Representation}

We  process and represent inputs from different modalities into sequential embeddings compatible with LLMs. Specifically, given an RGB image $I \in \mathbb{R}^{H \times W \times 3}$, where $H$ and $W$ denote the original resolution, we use a pre-trained visual encoder followed by an MLP-based connection module to extract and project image features into the LLM embedding space as:
$X_{I} = \text{MLP}(\text{Encoder}(I))$, resulting in a sequence representation of image tokens $X_{I} = [x_{1}^\mathcal{I}, \dots, x_{m}^\mathcal{I}] \in \mathbb{R}^{ m \times d}$, where $m = h \times w$ represents the number of image tokens. The MLP module ensures dimensional alignment with the LLMs' embedding space for seamless integration. For textual inputs $T \in \mathbb{Z}^{L}$, we tokenize and embed them to obtain a sequence of textual embeddings $X_{T} = [x_{1}^\mathcal{T}, \dots, x_{n}^\mathcal{T}] \in \mathbb{R}^{n \times d}$, where $n$ denotes the number of text tokens. The final multi-modal input representation $X$ is obtained by concatenating the visual and textual embeddings: $X= [X_{I}, X_{T}] \in \mathbb{R}^{(m+n) \times d}$, which is then fed into the LLM for joint processing of multi-modal information.

\subsubsection{Sparse MoIIE Forward}
The MoIIE is integrated into the LLM to facilitate efficient and adaptive multi-modal learning. The expert modules comprise intra-modality and inter-modality expert groups, formally defined as $\mathcal{E} = [E_1^\mathcal{T}, \dots, E_L^\mathcal{T}, E_{L+1}^\mathcal{I}, \dots, E_{3L}^\mathcal{I}, E_{3L+1}^\mathcal{S}, \dots, E_{4L}^\mathcal{S}]$, where intra-modality expert groups $E^\mathcal{T}$ and $E^\mathcal{I}$ respectively process text and image tokens for learning modality-specific features. In contrast, the inter-modality expert group $E^\mathcal{S}$ processes tokens from both modalities to capture rich cross-modal associations. To balance modality-specific and shared representations, we establish the following expert allocation constraints:
$|E^\mathcal{T}|+|E^\mathcal{I}|=|E^\mathcal{S}|=2L$ and $|E^\mathcal{T}|=|E^\mathcal{I}|=L$. This configuration ensures that the number of intra-modality experts equals the number of inter-modality experts, while maintaining equal allocation to each modality's experts.

The token routing mechanism is governed by modality-specific routers $G^{\mathcal{M}}(x)$, which include a visual router $G^{\mathcal{I}}(x)$ and a textual router $G^{\mathcal{T}}(x)$. These routers dynamically assign each input token to its top-$K$ most relevant experts based on predicted activation probabilities as follows:
\begin{equation}
G^{\mathcal{M}}(x_{n}) = \text{Softmax}(\text{top-}K(x_{n} \cdot W_g^{\mathcal{M}})),
\end{equation}
where $x_{n} \in X$ is the representation of each token from the input sequence.$W_g^{\mathcal{M}}$ is the weight matrix used to compute the routing probabilities of tokens to experts. The sparse MoIIE forward computation is then formulated as:
\begin{equation}
\scalebox{0.75}{$
\text{MoIIE}(x_{n}) =
\begin{cases}
\sum_{i=1}^{K} G_{i}^{\mathcal{T}}(x_{n}) \cdot \text{E}_i^{\ \mathcal{T} / S}(x_{n}), & \text{if } x_{n} \in \mathcal{T}  \\
\sum_{i=1}^{K} G_{i}^{\mathcal{I}}(x_{n}) \cdot \text{E}_i^{\ \mathcal{I} / S}(x_{n}), & \text{if } x_{n} \in \mathcal{I}
\end{cases}
$}
\end{equation}
where each token $x_{n}$ is processed by its top-$K$ activated experts $\text{E}_i(x_{n})$, with their outputs aggregated through a probability-weighted summation. Specifically, image tokens are routed by the visual router $G^{\mathcal{I}}(x)$ to the top-$K$ experts selected from the combined pool of intra-vision and inter-modality groups $E^\mathcal{I / S}$, while textual tokens are routed by the textual router $G^{\mathcal{T}}(x)$ to the top-$K$ experts from the intra-language and inter-modality groups $E^\mathcal{T / S}$. The routing mechanism enables flexible expert selection patterns: each token may activate only intra-modality experts, only inter-modality experts, or a combination of both. This hierarchical routing strategy ensures efficient multi-modal learning while simultaneously modeling both modality-specific features and cross-modal associations. All experts in $\mathcal{E}$ are initialized from the LLM’s FFN layers, preserving prior knowledge while enabling adaptive multi-modal specialization.

\subsection{Training Recipe}
\label{section:training}
To balance simplicity and efficacy, we propose a two-stage training strategy that effectively activates the capabilities of the MoIIE architecture. In the first stage, we focus on pretraining the connection module to align visual representations with the LLM’s linguistic embedding space, enabling consistent processing of inputs across different modalities. 
In the second stage, we initialize both intra-modality and inter-modality experts using the pre-trained FFNs from the base LLM and fine-tune the entire model with diverse visual instruction datasets. During this stage, the training objectives include multi-modal expert learning for capturing modality-specific features and cross-modal associations, and optimizing versatile multi-modal understanding and instruction-following capability.


\paragraph{Training Objective} To ensure load balancing among experts within the MoIIE module while optimizing the model for overall multi-modal understanding, the training objective $\mathcal{L}$ is composed of two components: the language modeling cross-entropy loss  $\mathcal{L}_{\text{lm}}$ and an auxiliary loss $\mathcal{L}_{\text{aux}}$ aimed at load balancing among experts~\cite{Mixtral}. 
\begin{equation}
    \mathcal{L} = \mathcal{L}_{\text{lm}} + \alpha \cdot \mathcal{L}_{\text{aux}},
\end{equation}
where $\alpha$ is a weighting coefficient set to 0.001. Notably, higher values of $\alpha$ adversely impact model performance, as detailed in Section \ref{sec:load balance}. The auxiliary loss $\mathcal{L}_{\text{aux}}$ is defined as:
\begin{equation}
\mathcal{L}_{\text{aux}} = |\mathcal{E}| \cdot \sum_{i=1}^{|\mathcal{E}|} \left( \mathbb{E}_x \left[ G_i(x) \right] \cdot \mathbb{E}_x \left[ \mathbf{1}_i(x) \right] \right),
\end{equation}
\( G_i(x) \) is the routing probability of assigning token \( x \) to expert \( E_i \), and \( \mathbf{1}_i(x) \) denotes the indicator function that equals 1 when expert $E_i$ is activated for token $x$ and 0 otherwise.

\section{Experiments}
\label{section:Exp}
\subsection{Implementation Details}
We employ a pre-trained SigLIP model~\cite{zhai2023sigmoidlosslanguageimage} and CLIP~\cite{radford2021clip}in Appendix~\ref{sec:More Experimental Results} as visual encoder, a two-layer MLP as connection module. For the LLM backbone, we utilize phi-3-mini~\cite{abdin2024phi3technicalreporthighly} and LLaMA3-8b~\cite{llama3modelcard}. Our method and all compared MoE architectures use a 4-experts configuration. Specifically, our MoIIE module includes 2 intra-modality experts (one for vision and one for language) and 2 inter-modality experts. More detailed discussion of expert settings is provided in the ablation studies~\ref{sec:Ablation Studies}.

\begin{table*}[!th]
\centering
\caption{\textbf{Comparison results between MoIIE and other ablated architectural variants across comprehensive multi-modal benchmarks with pre-trained SigLIP-So400m-384 as visual encoder} “Data” indicates the amount of visual instruction data. SEED-I, HalluB and MMMU$_\text{v}$ are abbreviations for SEED-Image, HallusionBench, and the MMMU validation subset, respectively. Bold numbers represent the best performance in each column.}
\setlength\tabcolsep{2pt}
\resizebox{1\textwidth}{!}{
\begin{tabular}{ccccccccccccccccc}
\toprule
\multicolumn{1}{c|}{} & \multicolumn{1}{c|}{} & \multicolumn{1}{c|}{} &
MMBench & GQA & VQAv2 & MMVet & \multicolumn{1}{c|}{SEED-I} &
POPE & \multicolumn{1}{c|}{HalluB} &
TextVQA & DocVQA & \multicolumn{1}{c|}{ChartQA} &
AI2D & MMMU$_\text{v}$ & \multicolumn{1}{c|}{Mathvista} & \\
\multicolumn{1}{c|}{\multirow{-2}{*}{}} &
\multicolumn{1}{c|}{\multirow{-2}{*}{Data}} &
\multicolumn{1}{c|}{\multirow{-2}{*}{Backbone}} &
\multicolumn{5}{c|}{General Multi-modal QA} &
\multicolumn{2}{c|}{Hallucination} &
\multicolumn{3}{c|}{OCR-based QA} &
\multicolumn{3}{c|}{Knowledge-based QA} &
\multirow{-2}{*}{AVG} \\
\midrule
\multicolumn{1}{c|}{Dense} 
  & \multicolumn{1}{c|}{2M} 
  & \multicolumn{1}{c|}{Phi-3-mini} 
  & 74.4 & 62.7 & 81.6 & \textbf{42.9} 
  & \multicolumn{1}{c|}{70.2} & 86.2 & \multicolumn{1}{c|}{28.3} 
  & 65.1 & 41.9 & \multicolumn{1}{c|}{53.2} 
  & 64.3 & 41.3 & \multicolumn{1}{c|}{30.8} 
  & 57.1 \\
\rowcolor[HTML]{FFFFFF}
\multicolumn{1}{c|}{Vanilla MoE} 
  & \multicolumn{1}{c|}{2M} 
  & \multicolumn{1}{c|}{Phi-3-mini} 
  & 75.1 & 62.8 & 81.5 & 42.8 
  & \multicolumn{1}{c|}{70.6} & 86.3 & \multicolumn{1}{c|}{28.8} 
  & 65.1 & 42.2 & \multicolumn{1}{c|}{53.1} 
  & 65.2 & 40.6 & \multicolumn{1}{c|}{29.9} 
  & 57.2 \\
\rowcolor[HTML]{FFFFFF}
\multicolumn{1}{c|}{Modality MoE} 
  & \multicolumn{1}{c|}{2M} 
  & \multicolumn{1}{c|}{Phi-3-mini} 
  & 75.1 & 62.9 & 81.7 & 41.0 
  & \multicolumn{1}{c|}{71.0} & 86.2 & \multicolumn{1}{c|}{29.3} 
  & \textbf{66.0} & 42.7 & \multicolumn{1}{c|}{53.1} 
  & 65.5 & 40.4 & \multicolumn{1}{c|}{31.0} 
  & 57.4 \\
\rowcolor[HTML]{E1EAFF}
\multicolumn{1}{c|}{MoIIE}            
  & \multicolumn{1}{c|}{2M}   
  & \multicolumn{1}{c|}{Phi-3-mini} 
  & \textbf{75.3} & \textbf{63.2} & \textbf{81.8} & 42.8    
  & \multicolumn{1}{c|}{\textbf{71.2}} & \textbf{86.6} & \multicolumn{1}{c|}{\textbf{30.5}} 
  & 65.4 & \textbf{42.9} & \multicolumn{1}{c|}{\textbf{53.3}} 
  & \textbf{65.9} & \textbf{41.3} & \multicolumn{1}{c|}{\textbf{31.9}} 
  & \textbf{57.9} \\ 
\midrule
\rowcolor[HTML]{DEE0E3}
\multicolumn{17}{c}{\cellcolor[HTML]{DEE0E3}With More Visual Instruction Data} \\ 
\midrule
\multicolumn{1}{c|}{Dense} 
  & \multicolumn{1}{c|}{2.7M} 
  & \multicolumn{1}{c|}{Phi-3-mini} 
  & 75.0 & 63.4 & 81.6 & 41.1 
  & \multicolumn{1}{c|}{70.1} & 87.0 & \multicolumn{1}{c|}{31.7} 
  & 64.6 & 48.2 & \multicolumn{1}{c|}{57.5} 
  & 73.7 & 40.1 & \multicolumn{1}{c|}{31.0} 
  & 58.8 \\
\multicolumn{1}{c|}{Vanilla MoE} 
  & \multicolumn{1}{c|}{2.7M} 
  & \multicolumn{1}{c|}{Phi-3-mini} 
  & 75.2 & 63.3 & 81.4 & 42.2 
  & \multicolumn{1}{c|}{70.6} & 87.2 & \multicolumn{1}{c|}{30.8} 
  & 65.2 & 47.8 & \multicolumn{1}{c|}{57.2} 
  & 74.7 & 42.2 & \multicolumn{1}{c|}{31.2} 
  & 59.1 \\
\multicolumn{1}{c|}{Modality MoE} 
  & \multicolumn{1}{c|}{2.7M} 
  & \multicolumn{1}{c|}{Phi-3-mini} 
  & 73.9 & 63.5 & 81.7 & 40.1 
  & \multicolumn{1}{c|}{70.4} & 87.1 & \multicolumn{1}{c|}{\textbf{32.9}} 
  & 65.7 & \textbf{48.9} & \multicolumn{1}{c|}{58.8} 
  & 74.1 & 40.4 & \multicolumn{1}{c|}{31.3} 
  & 59.1 \\
\rowcolor[HTML]{E1EAFF}
\multicolumn{1}{c|}{MoIIE}            
  & \multicolumn{1}{c|}{2.7M} 
  & \multicolumn{1}{c|}{Phi-3-mini} 
  & \textbf{75.9} & \textbf{64.1} & \textbf{82.0} & \textbf{42.7}     
  & \multicolumn{1}{c|}{\textbf{71.8}} & \textbf{87.2} & \multicolumn{1}{c|}{30.7} 
  & \textbf{66.9} & 48.5 & \multicolumn{1}{c|}{\textbf{58.8}} 
  & \textbf{75.9} & \textbf{42.4} & \multicolumn{1}{c|}{\textbf{32.5}} 
  & \textbf{60.0} \\ 
\midrule
\rowcolor[HTML]{DEE0E3}
\multicolumn{17}{c}{\cellcolor[HTML]{DEE0E3}With Larger LLM Backbone} \\ 
\midrule
\multicolumn{1}{c|}{Dense} 
  & \multicolumn{1}{c|}{2M} 
  & \multicolumn{1}{c|}{LLaMA3-8B} 
  & 74.8 & 64.8 & 82.2 & \textbf{47.9} 
  & \multicolumn{1}{c|}{72.1} & 86.6 & \multicolumn{1}{c|}{30.1} 
  & 67.2 & 43.4 & \multicolumn{1}{c|}{54.0} 
  & 76.6 & 41.1 & \multicolumn{1}{c|}{30.9} 
  & 59.4 \\
\multicolumn{1}{c|}{Vanilla MoE} 
  & \multicolumn{1}{c|}{2M} 
  & \multicolumn{1}{c|}{LLaMA3-8B} 
  & 75.5 & 64.5 & 82.1 & 46.0 
  & \multicolumn{1}{c|}{71.6} & 86.3 & \multicolumn{1}{c|}{33.0} 
  & 67.3 & 43.8 & \multicolumn{1}{c|}{54.7} 
  & 76.2 & 40.7 & \multicolumn{1}{c|}{30.6} 
  & 59.4 \\
\multicolumn{1}{c|}{Modality MoE} 
  & \multicolumn{1}{c|}{2M} 
  & \multicolumn{1}{c|}{LLaMA3-8B} 
  & 75.6 & 64.3 & 82.3 & 46.3 
  & \multicolumn{1}{c|}{72.3} & 86.7 & \multicolumn{1}{c|}{35.0} 
  & 67.6 & \textbf{44.9} & \multicolumn{1}{c|}{55.4} 
  & 76.7 & 41.6 & \multicolumn{1}{c|}{31.2} 
  & 60.0 \\
\rowcolor[HTML]{E1EAFF}
\multicolumn{1}{c|}{MoIIE}            
  & \multicolumn{1}{c|}{2M}   
  & \multicolumn{1}{c|}{LLaMA3-8B} 
  & \textbf{75.7} & \textbf{64.9} & \textbf{82.3} & 47.5    
  & \multicolumn{1}{c|}{\textbf{73.0}} & \textbf{87.0} & \multicolumn{1}{c|}{\textbf{36.5}} 
  & \textbf{67.9} & 44.7 & \multicolumn{1}{c|}{\textbf{56.0}} 
  & \textbf{76.9} & \textbf{42.8} & \multicolumn{1}{c|}{\textbf{32.2}} 
  & \textbf{60.6} \\ 
\bottomrule
\end{tabular}}
\label{table:model arch}
\end{table*}

During training, we utilize the AdamW optimizer~\cite{kingma2017adammethodstochasticoptimization} with a cosine learning rate scheduler for one epoch across both stages. In the first stage, only the connection module is optimized with a learning rate of \(1 \times 10^{-3}\), using the Bunny-pretrain-LAION-2M dataset~\cite{he2024efficientmultimodallearningdatacentric}. In the second stage, all parameters are unfrozen for joint SFT and sparse upcycling training. The learning rate is set to \(2 \times 10^{-6}\) for the visual encoder and \(2 \times 10^{-5}\) for all other components. For this stage, we attempt training with different scales of visual instruction data: (1) 1.3M samples from MGM-Instruction~\cite{li2024minigeminiminingpotentialmultimodality}; (2) 2M samples by adding Bunny-695K~\cite{he2024efficientmultimodallearningdatacentric}; (3) 2.7M samples by further incorporating LLaVA-NEXT-779k~\cite{liu2024llava}(4)8M advanced training data from LLaVA-OV~\cite{li2024llavaonevisioneasyvisualtask}. Training is conducted using DeepSpeed~\cite{song2023deepspeed4scienceinitiativeenablinglargescale} with ZeRO-3 optimization.  All compared baselines in the main results and our method follow the same training setup and data configurations. Further implementation details and results are provided in Appendix~\ref{section:appendix}.


\subsection{Evaluation Benchmarks}
We conduct a comprehensive evaluation across 13 diverse multi-modal benchmarks. Specifically, general multi-modal benchmarks include MMBench-EN~\cite{liu2024mmbenchmultimodalmodelallaround}, MM-Vet~\cite{yu2024mmvetevaluatinglargemultimodal}, GQA~\cite{hudson2019gqanewdatasetrealworld}, VQAv2 and SEED-Image ~\cite{li2023seedbenchbenchmarkingmultimodalllms}. For knowledge-based question answering, we utilize MMMU validation subset~\cite{yue2024mmmumassivemultidisciplinemultimodal}, AI2D~\cite{kembhavi2016diagramworthdozenimages},SciQA-IMG~\cite{lu2022learnexplainmultimodalreasoning} and Mathvista~\cite{lu2024mathvistaevaluatingmathematicalreasoning}. For OCR-based question answering, we assess performance on TextVQA~\cite{singh2019vqamodelsread}, ChartQA~\cite{masry2022chartqabenchmarkquestionanswering}, and  DocVQA~\cite{mathew2021docvqadatasetvqadocument}. Additionally, we evaluate hallucination robustness using POPE~\cite{li2023evaluatingobjecthallucinationlarge} and HallusionBench~\cite{guan2024hallusionbenchadvanceddiagnosticsuite}.

\subsection{Main Results}
\label{sec:scaling}
We compare our MoIIE-equipped model against three baseline variants: the original dense model (Dense) without MoE, the vanilla MoE model~\cite{Mixtral} that routes tokens from all modalities to a shared pool of experts (Vanilla MoE), and the model with modality-specific experts where text and image tokens are routed to their respective specialized experts (Modality MoE). The comparisons are conducted across multiple visual instruction tuning data scales and varying LLM backbones, as shown in Table~\ref{table:model arch}. The key findings are as follows:

(1) The MoIIE module consistently outperforms all other architectural variants across most benchmarks with different data scales, visual encoders, and LLM backbone sizes, demonstrating the powerful generalization capability of our MoIIE. Integrating both modality-specific features and cross-modal associations can effectively enhance performance on diverse multi-modal tasks. Further analysis of expert modules is provided in Appendix~\ref{sec:Experts Analysis}.

(2) Specifically, our MoIIE module is particularly effective on knowledge-based QA and hallucination benchmarks that requiring sophisticated cross-modal interaction between textual concepts and corresponding visual entity regions. In contrast, OCR-based QA tasks primarily involve interpreting information verbatim from images, where Modality MoE can already achieves competitive performance. This performance difference highlights MoIIE's capacity to handle both modality-specific features and cross-modal associations.

\begin{figure*}[!th]
    \centering
    \includegraphics[width=1.0\linewidth]{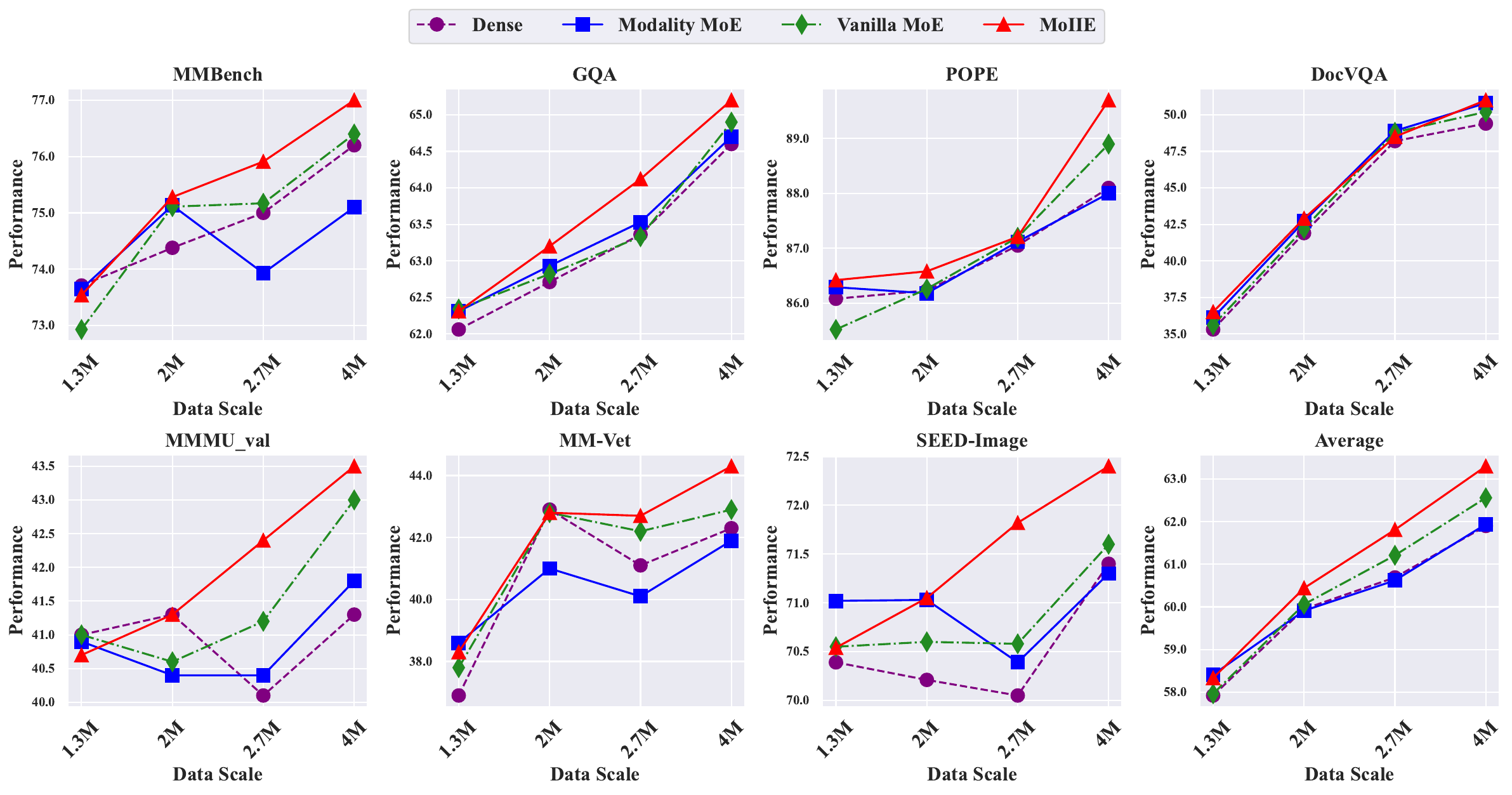}
    \caption{\textbf{Performance variation across different visual instruction tuning data scales }. MoIIE outperforms other architectural variants in achieving superior scaling efficiency.}
    \label{fig:data_scale}
\end{figure*}

(3) We further provide detailed performance trends with increasing fine-tuning data in Figure~\ref{fig:data_scale}.
As the data scale continues to increase, our MoIIE progressively improves while other architectures, especially Dense and Modality MoE, encounter performance limitations.
Moreover, with larger training datasets, our MoIIE consistently outperforms all alternatives, with the performance gap between MoIIE and other architectures widening. This suggests that the MoIIE framework offers superior scaling properties for multi-modal learning, effectively leveraging larger datasets to enhance representation power without the parameter inefficiency of dense models or the limited cross-modal reasoning of strictly modality-separated experts. 



\subsection{Comparison with MoE-based LLMs Initialized LVLMs}
We compare our MoIIE, pre-trained on the LLaVA-OV-Single-Image dataset~\cite{li2024llavaonevisioneasyvisualtask}, with open-source LVLMs initialized from MoE-based LLMs. Models such as SPHINX-X~\cite{liu2024sphinxxscalingdataparameters}, MGM~\cite{li2024minigeminiminingpotentialmultimodality}, and CuMO~\cite{li2024cumoscalingmultimodalllm}, built on Mixtral-8x7B~\cite{Mixtral}, exploit large-scale language data for robust initialization but require substantial computational resources, limiting practicality in constrained settings. By contrast, MoIIE attains comparable performance across diverse multimodal benchmarks with 17\% fewer activated parameters, thereby reducing both training and deployment costs. Moreover, it enables constructing MoE-based LVLMs from arbitrary LLM backbones, offering flexible expert configurations without performance degradation.

\begin{table*}[!th]
\centering
\caption{\textbf{Comparisons of MoIIE with open-source LVLMs utilizing pre-trained MoE-based LLM backbones.} In addition to performance gains, our MoIIE offers the flexibility to be constructed using any dense LLM backbone. ``Act. Param'' indicates activated parameters while ``All Param'' indicates the total number of model parameters.}
\setlength\tabcolsep{3pt}
\resizebox{0.8\textwidth}{!}{
\begin{tabular}{cccccccccccc}
\toprule
\multicolumn{1}{c|}{model}                             & \multicolumn{1}{c|}{Act. Param}                     & \multicolumn{1}{c|}{All Param}                        & TextVQA                     & GQA                         & VQAv2                       & POPE                                 & MMBench                     & MMVet                                & MMMU$_\text{v}$                          & Mathvista                   & SciQA-IMG                   \\ \midrule
\rowcolor[HTML]{DEE0E3} 
\multicolumn{12}{c}{\cellcolor[HTML]{DEE0E3}MoE-based LVLMs model with pre-trained MoE LLM backbone}                                                                                                                                                                                                                                                                                                                                                                       \\ \midrule
\multicolumn{1}{c|}{{\color[HTML]{1F2329} SPHINX-X~\cite{liu2024sphinxxscalingdataparameters}}}   & \multicolumn{1}{c|}{{\color[HTML]{1F2329} -}}      & \multicolumn{1}{c|}{{\color[HTML]{1F2329} $\sim$40B}} & {\color[HTML]{1F2329} 68.0}   & {\color[HTML]{1F2329} 63.8} & {\color[HTML]{1F2329} 81.1} & {\color[HTML]{1F2329} \textbf{89.6}} & {\color[HTML]{1F2329} 71.3} & {\color[HTML]{1F2329} 40.9}          & {\color[HTML]{1F2329} 31.1}        & {\color[HTML]{1F2329} 42.7} & {\color[HTML]{1F2329} 74.5} \\
\multicolumn{1}{c|}{{\color[HTML]{1F2329} MGM~\cite{li2024minigeminiminingpotentialmultimodality}}}        & \multicolumn{1}{c|}{{\color[HTML]{1F2329} 13.5B}}  & \multicolumn{1}{c|}{{\color[HTML]{1F2329} $\sim$40B}} & {\color[HTML]{1F2329} 69.2} & {\color[HTML]{1F2329} -}    & {\color[HTML]{1F2329} -}    & {\color[HTML]{1F2329} -}             & {\color[HTML]{1F2329} 75.6} & {\color[HTML]{1F2329} 45.8}          & {\color[HTML]{1F2329} 41.8}        & {\color[HTML]{1F2329} 41.8} & {\color[HTML]{1F2329} -}    \\
\multicolumn{1}{c|}{{\color[HTML]{1F2329} CuMO~\cite{li2024cumoscalingmultimodalllm}}}       & \multicolumn{1}{c|}{{\color[HTML]{1F2329} 13.5B}}  & \multicolumn{1}{c|}{{\color[HTML]{1F2329} $\sim$40B}} & {\color[HTML]{1F2329} 66.0}   & {\color[HTML]{1F2329} 63.8} & {\color[HTML]{1F2329} 81.8} & {\color[HTML]{1F2329} 85.7}          & {\color[HTML]{1F2329} 75.3} & {\color[HTML]{1F2329} \textbf{48.7}} & {\color[HTML]{1F2329} \textbf{45.0}} & {\color[HTML]{1F2329} 38.2} & {\color[HTML]{1F2329} 77.9} \\ \midrule
\rowcolor[HTML]{DEE0E3} 
\multicolumn{12}{c}{\cellcolor[HTML]{DEE0E3}MoE-based LVLMs model from dense LLM backbone}                                                                                                                                                                                                                                                                                                                                                                                 \\ \midrule
\rowcolor[HTML]{E1EAFF} 
\multicolumn{1}{c|}{\cellcolor[HTML]{E1EAFF}MoIIE-A5B}  & \multicolumn{1}{c|}{\cellcolor[HTML]{E1EAFF}5.5B}  & \multicolumn{1}{c|}{\cellcolor[HTML]{E1EAFF}7B}       & 69.5                        & 63.4                        & 82.5                        & 87.1                                 & \textbf{77.8}               & 42.6                                 & 43.6                               & \textbf{46.4}               & \textbf{90.1}               \\
\rowcolor[HTML]{E1EAFF} 
\multicolumn{1}{c|}{\cellcolor[HTML]{E1EAFF}MoIIE-A11B} & \multicolumn{1}{c|}{\cellcolor[HTML]{E1EAFF}11.3B} & \multicolumn{1}{c|}{\cellcolor[HTML]{E1EAFF}16B}      & \textbf{70.5}               & \textbf{64.5}               & \textbf{82.9}               & 87.3                                 & 77.2                        & 47.3                                 & 42.7                               & 46.0                          & 89.4                        \\ \bottomrule
\end{tabular}}
\label{table:benchmark}
\end{table*}

\subsection{The Discussion of Training Pipeline}
Converting dense LLMs into MoE-based LVLMs typically follows a three-stage pipeline~\cite{lin2024moellavamixtureexpertslarge,chen2024llavamolesparsemixturelora}: (1) multimodal pretraining, (2) supervised fine-tuning (SFT) with multimodal instructions, and (3) sparse upcycling to obtain MoE-based LVLMs. This process is complex, and limited training in the final stage often weakens generalization. We propose a simplified two-stage approach: after pretraining, SFT and sparsification are combined into one phase, enabling direct activation of both multimodal and MoE features. Specifically, the three-stage pipeline uses 2M visual instruction samples for SFT (tuning all parameters) and LLaVA-NEXT-779k for sparse upcycling (only tuning MoE layers), while our two-stage method uses 2.7M samples for joint SFT and sparse upcycling (tuning all parameters). As shown in Table~\ref{training_stage}, our approach consistently outperforms the three-stage pipeline under the same conditions, offering both simplicity and improved effectiveness.

\begin{table*}[th!]
\centering
\caption{\textbf{Comparisons of different training strategies.} Pretrain-SFT-MoE represents the three-stage training method where MoE denotes sparse upcycling from multi-modal dense checkpoints, tuning only MoE layers~\cite{lin2024moellavamixtureexpertslarge} after SFT. Pretrain-SFT\&MoE is our proposed two-stage training method, in which sparse upcycling is integrated with SFT, tuning the full model parameters simultaneously.}
\setlength\tabcolsep{2.2pt}
\resizebox{0.8\textwidth}{!}{
\begin{tabular}{c|c|ccccc|cc|ccc|ccc|c}
\toprule
\multicolumn{1}{c|}{} 
  & \multicolumn{1}{c|}{} 
  & MMBench & GQA & VQAv2 & MMVet & \multicolumn{1}{c|}{SEED-I} 
  & POPE & \multicolumn{1}{c|}{HalluB} 
  & TextVQA & DocVQA & \multicolumn{1}{c|}{ChartQA} 
  & AI2D & MMMU$_\text{v}$ & \multicolumn{1}{c|}{Mathvista} 
  & \\
\multicolumn{1}{c|}{\multirow{-2}{*}{Method}} 
  & \multicolumn{1}{c|}{\multirow{-2}{*}{Stage}} 
  & \multicolumn{5}{c|}{General Multi-modal QA} 
  & \multicolumn{2}{c|}{Hallucination} 
  & \multicolumn{3}{c|}{OCR-based QA} 
  & \multicolumn{3}{c|}{Knowledge-based QA} 
  & \multirow{-2}{*}{AVG} \\
\midrule
Pretrain-SFT-MoE       
  & 3  
  & 75.6 & 63.1 & 81.6 & 41.2 & 70.3  
  & 86.9 & 31.0  
  & 64.8 & 48.2 & 56.2  
  & 63.8 & 39.3 & 31.3  
  & 57.9 \\
\rowcolor[HTML]{E1EAFF}
Pretrain-SFT\&MoE    
  & 2  
  & \textbf{75.7} & \textbf{63.5} & \textbf{81.8} & \textbf{42.7} & \textbf{71.2}  
  & \textbf{87.2} & \textbf{32.1}  
  & \textbf{65.3} & \textbf{49.6} & \textbf{58.8}  
  & \textbf{65.3} & \textbf{41.9} & \textbf{32.5}  
  & \textbf{59.0} \\ 
\midrule
\end{tabular}}
\label{training_stage}
\end{table*}

\subsection{Ablation Study}
\label{sec:Ablation Studies}

\begin{table*}[th!]
\centering
\caption{\textbf{Ablation study for various MoIIE module configurations.} This table compares the performance of different MoIIE setups, including variations in the MoIIE location, load balance coefficient, number of experts and experts settings. 
}
\setlength\tabcolsep{2.5pt}
\resizebox{0.8\textwidth}{!}{
\begin{tabular}{c|c|c|cccccccc}
\toprule
Ablated Aspects & Original & Ablated Setting & TextVQA & MMBench & GQA & MMMU$_\text{v}$ & POPE & MMVet & SEED-Image & AVG \\ \midrule
Number of Experts & 4 & 8 & 65.1 & 75.6 & 63.4 & 42.2 & 86.5 & 43.0 & 71.2 & 63.9(+0.2) \\ \midrule
Experts Balance(8 experts) & Balance & Unbalance & 65.4 & 75.0 & 63.4 & 41.1 & 86.8 & 41.3 & 71.6 & 63.5(-0.2) \\ \midrule
MoIIE Module Location & Interleaved & Full & 65.8 & 75.8 & 63.1 & 40.3 & 87.0 & 42.6 & 70.6 & 63.6(-0.1) \\ \midrule
Load Balance Coefficient & 0.001 & 0.01 & 65.5 & 75.4 & 63.2 & 39.0 & 86.4 & 42.0 & 70.7 & 63.2(-0.5) \\ \midrule
\rowcolor[HTML]{E1EAFF} 
Ours(MoIIE-A5B) & - & - & 65.4 & 75.3 & 63.2 & 41.3 & 86.6 & 42.8 & 71.1 & 63.7 \\ \midrule
\end{tabular}}
\label{table:ablation}
\end{table*}


\paragraph{The Impact of the Number of Experts}
Increasing the number of experts generally enhances MoE performance~\cite{lepikhin2020gshardscalinggiantmodels,fedus2022switch}. As shown in Table~\ref{table:ablation}, expanding the MoIIE expert pool from 4 to 8 yields gains of 0.3, 0.2, and 0.9 points on MMBench, GQA, and MMMU, respectively, while results on POPE and TextVQA remain stable or slightly decline. These findings suggest that larger expert pools improve generalization on complex benchmarks but may require more data to fully realize their capacity.

\paragraph{The Impact of Experts Balance}
Maintaining a balanced allocation of intra- and inter-modality experts is crucial for performance. As shown in Table~\ref{table:ablation}, under 8-expert setup, Balanced configuration (2 vision, 2 language, 4 shared; 1:1 ratio) consistently outperforms the Unbalanced configuration (3 vision, 3 language, 2 shared; 3:1 ratio). The balanced design ensures more effective cross-modal interaction, getting stronger overall performance.

\paragraph{Layer Location of MoIIE module}
We investigate two configurations for integrating MoIIE into LVLMs: an interleaved design, where MoIIE modules are inserted every other layer alongside dense layers, and a full design, where all layers are replaced with MoIIE modules~\cite{lin2024moellavamixtureexpertslarge,llama4modelcard}. As shown in Table~\ref{table:ablation}, the full design does not yield notable performance gains but significantly increases training cost and computational overhead due to the larger parameter count. By contrast, the interleaved configuration achieves comparable results with substantially higher efficiency and lower resource usage. We therefore adopt the interleaved design as the optimal balance of performance, efficiency, and scalability for MoIIE-based LVLMs.
\paragraph{Impact of Load Balance Coefficient}
\label{sec:load balance}
In MoE architectures, there is a trade-off between load balancing and performance. While the auxiliary loss promotes balanced expert usage~\cite{cai2024surveymixtureexperts}, excessive balancing can hinder multimodal learning and weaken cross-modal associations. As shown in Table~\ref{table:ablation}, increasing the auxiliary loss weight $\alpha$ consistently reduces performance, especially on MMMU\textsubscript{val} and MM-Vet. These results underscore the importance of tuning $\alpha$ to maintain expert specialization and robust multimodal understanding.

\subsection{Visualization of Token Pathways}
\begin{figure}[!ht]
    \centering
    \includegraphics[width=0.5\textwidth]{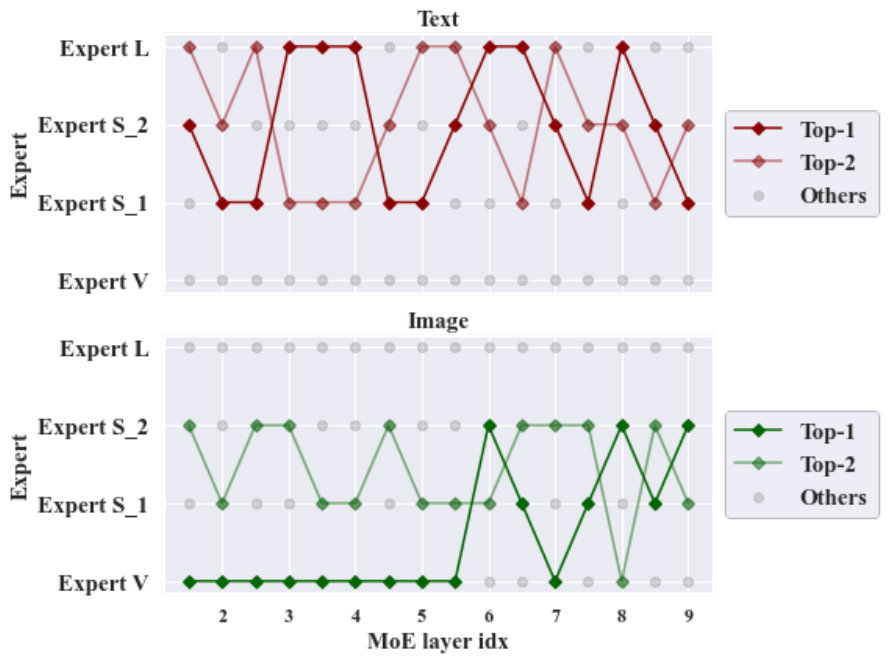}
    \caption{\textbf{Visualization of experts activated pathways.} The figure shows the top-2 activated experts for text and image tokens, with Expert V and Expert L are intra-modality experts, Expert S are inter-modality experts.}
    \label{fig:router_vis}
\end{figure}
As shown in Figure~\ref{fig:router_vis}, we visualize expert activation pathways for image and text tokens in the MoIIE module on the MME test set. Distinct layer-wise patterns emerge: in shallow layers, tokens primarily activate intra-modality experts (Expert V for vision, Expert L for text), reflecting a focus on modality-specific feature extraction with limited cross-modal interaction. In deeper layers, activation shifts toward inter-modality experts (Expert S), enabling stronger cross-modal fusion. Notably, both modalities often converge on the same inter-modality experts (e.g., Expert S\_1 in the 9th layer, Expert S\_2 in the 12th). These results demonstrate that MoIIE dynamically balances modality-specific and cross-modal processing, supporting early-stage representation learning and late-stage integration.

\section{Related Works}

\paragraph{Large Vision-Language Models}
Recent advances in LLMs~\cite{bai2023qwenlan,llama3modelcard,touvron2023llama2,vicuna2023,jiang2023mistral,bi2024deepseek,team2023internlm,openai2023gpt4,openai2023chatgpt,tou2023llama} have shown strong generalization and instruction-following abilities, spurring research on Large Vision-Language Models (LVLMs)~\cite{li2023blip,alayrac2022flamingovisuallanguagemodel,liu2023improved,liu2023visual,liu2024llava}. Progress in LVLMs has been driven by high-quality data~\cite{he2024efficientmultimodallearningdatacentric,chen2024allava,zhang2024llavarenhancedvisualinstruction}, extended training schemes~\cite{bai2023qwen,lu2024deepseekvlrealworldvisionlanguageunderstanding,laurençon2024buildingbetterunderstandingvisionlanguage}, support for high-resolution inputs~\cite{li2024monkeyimageresolutiontext,wang2024cogvlmvisualexpertpretrained}, and multi-encoder designs~\cite{shi2024eagleexploringdesignspace,liu2024sphinxxscalingdataparameters,fan2024mousipolyvisualexpertvisionlanguagemodels}. The latest open-source LVLMs~\cite{li2024llavaonevisioneasyvisualtask,wang2024qwen2vlenhancingvisionlanguagemodels,chen2025expandingperformanceboundariesopensource} achieve state-of-the-art results by integrating large-scale datasets, dynamic resolution, and advanced LLMs.

\paragraph{Mixture of Experts in LLMs}
The MoE paradigm~\cite{6797059} scales model capacity by activating only a subset of experts, balancing efficiency and performance. It typically replaces feed-forward layers with expert modules and relies on Top-K routing~\cite{lepikhin2020gshardscalinggiantmodels,du2022glamefficientscalinglanguage,fedus2022switch,zoph2022stmoedesigningstabletransferable,rajbhandari2022deepspeedmoeadvancingmixtureofexpertsinference,xue2024openmoeearlyeffortopen}. Sparse upcycling~\cite{komatsuzaki2023sparseupcyclingtrainingmixtureofexperts} further reduces costs by converting dense models into sparse ones. Recent MoE-based LLMs~\cite{xai2024grok,Mixtral,muennighoff2024olmoeopenmixtureofexpertslanguage,deepseekai2024deepseekv2strongeconomicalefficient,deepseekai2024deepseekv3technicalreport} improve stability and efficiency through large-scale training strategies.

\paragraph{Mixture of Experts in LVLMs}
MoE has also been applied to LVLMs, e.g., ARIA~\cite{li2025ariaopenmultimodalnative} and DeepSeek-VL2~\cite{wu2024deepseekvl2mixtureofexpertsvisionlanguagemodels}, which leverage MoE-based LLM backbones~\cite{Mixtral,deepseekai2024deepseekv2strongeconomicalefficient} but remain constrained by fixed expert settings and high training costs. Sparse upcycling~\cite{chen2024llavamolesparsemixturelora,lin2024moellavamixtureexpertslarge,lin2024momaefficientearlyfusionpretraining} improves scalability, yet struggles to capture both modality-specific and cross-modal interactions. More recent work~\cite{zhou2025enhancingmultimodalmodelsheterogeneous} extends parameter-efficient fine-tuning to multimodal experts, but the role of modality-specific expert groups remains underexplored. To address these gaps, we propose a modality-aware MoE architecture that integrates intra- and inter-modality experts. Combined with a two-stage training strategy, our approach enables flexible expert allocation and more scalable, high-performance LVLMs.

\vspace{0.1in}
\section{Conclusion}
\label{sec:conclusion}
In this study, we propose MoIIE, a novel Mixture-of-Experts architecture paired with a simple yet effective two-stage training strategy that enables the flexible construction of powerful MoE-based LVLMs from any dense LLM. MoIIE organizes experts into  specialized intra-modality and shared inter-modality groups to create dedicated pathways for both visual and linguistic processing while enabling rich cross-modal associations, significantly enhancing performance across a wide range of multi-modal tasks. Extensive experiments demonstrate that our framework achieves superior scaling efficiency and performance compared to existing dense and sparse LVLM architectures. 
\section*{Limitations} 
A primary limitation of our work lies in the relatively limited training data and its exclusive focus on vision-language modalities. In future work, we plan to address these limitations by incorporating more high-quality multi-task datasets, supporting dynamic input resolutions, and extending MoIIE to broader modalities such as speech, thereby advancing its generalization and applicability in real-world multi-modal scenarios.



\bibliography{main}

\appendix
\section{Appendix}
\label{section:appendix}

\begin{figure*}[th!]
\setlength{\abovecaptionskip}{0pt}
    \centering
    \includegraphics[width=0.8\linewidth]{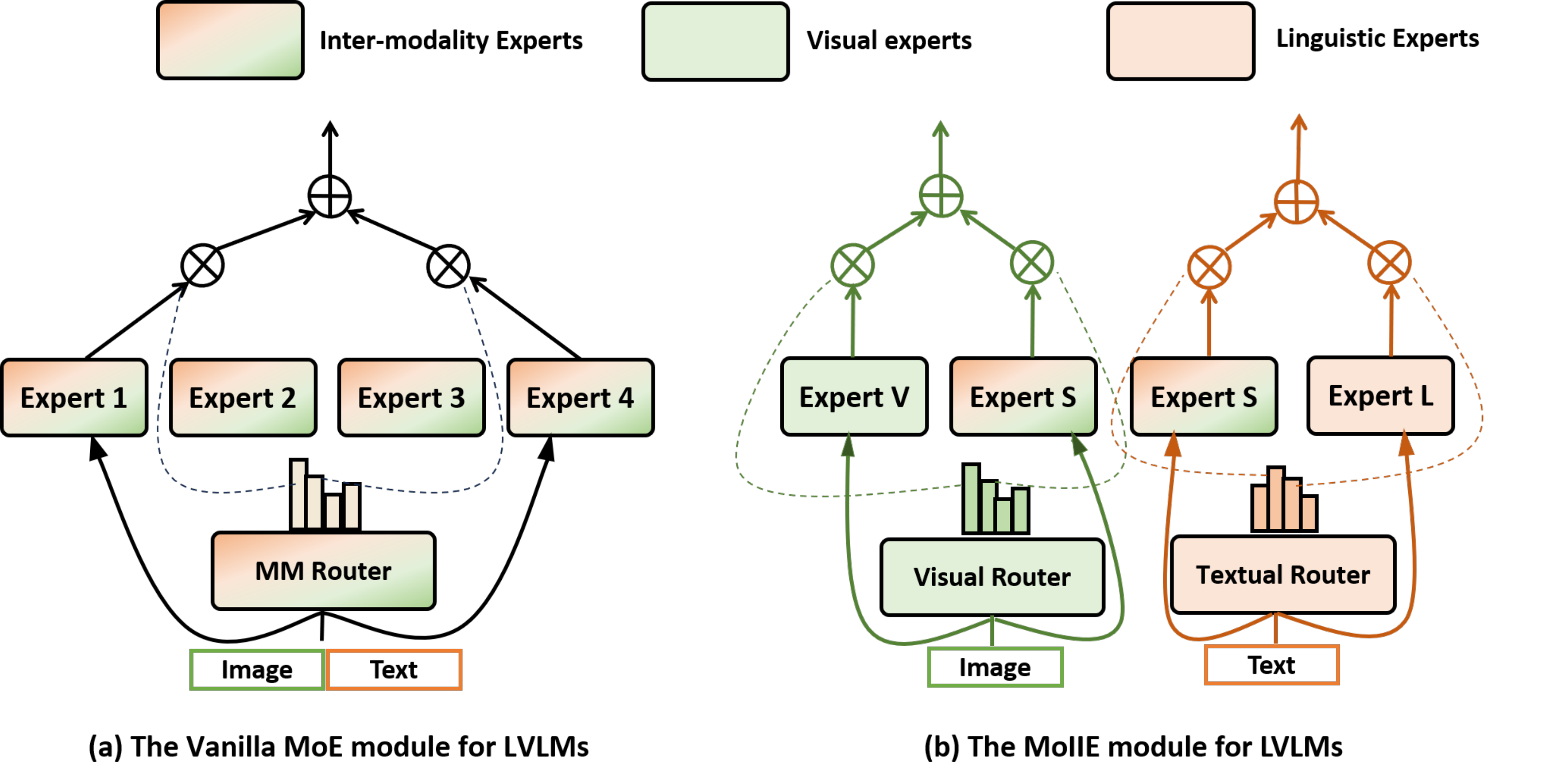}
    \caption{\textbf{Comparison between the Vanilla MoE and our MoIIE.}
\textbf{Left}: The vanilla MoE module~\cite{Mixtral} routes all modality tokens into a single group of experts.
\textbf{Right}: The MoIIE module introduces
intra-modality  and inter-modality experts group. Intra-modality Expert V for image tokens and Expert L for text tokens modeling modality specfic features. Inter-modality experts (Expert S) that process tokens from both modalities, modeling cross-modal associations, capable of handling tokens from both modalities. The visual router routes image tokens to
Expert V and Expert S, while the textual router routes text tokens to Expert L and Expert S.}
    \label{fig:arch_compare}
\end{figure*}



\subsection{Visualization of activated experts}
\begin{figure*}[h]
\setlength{\abovecaptionskip}{0pt}
    \centering
    \includegraphics[width=0.8\linewidth]{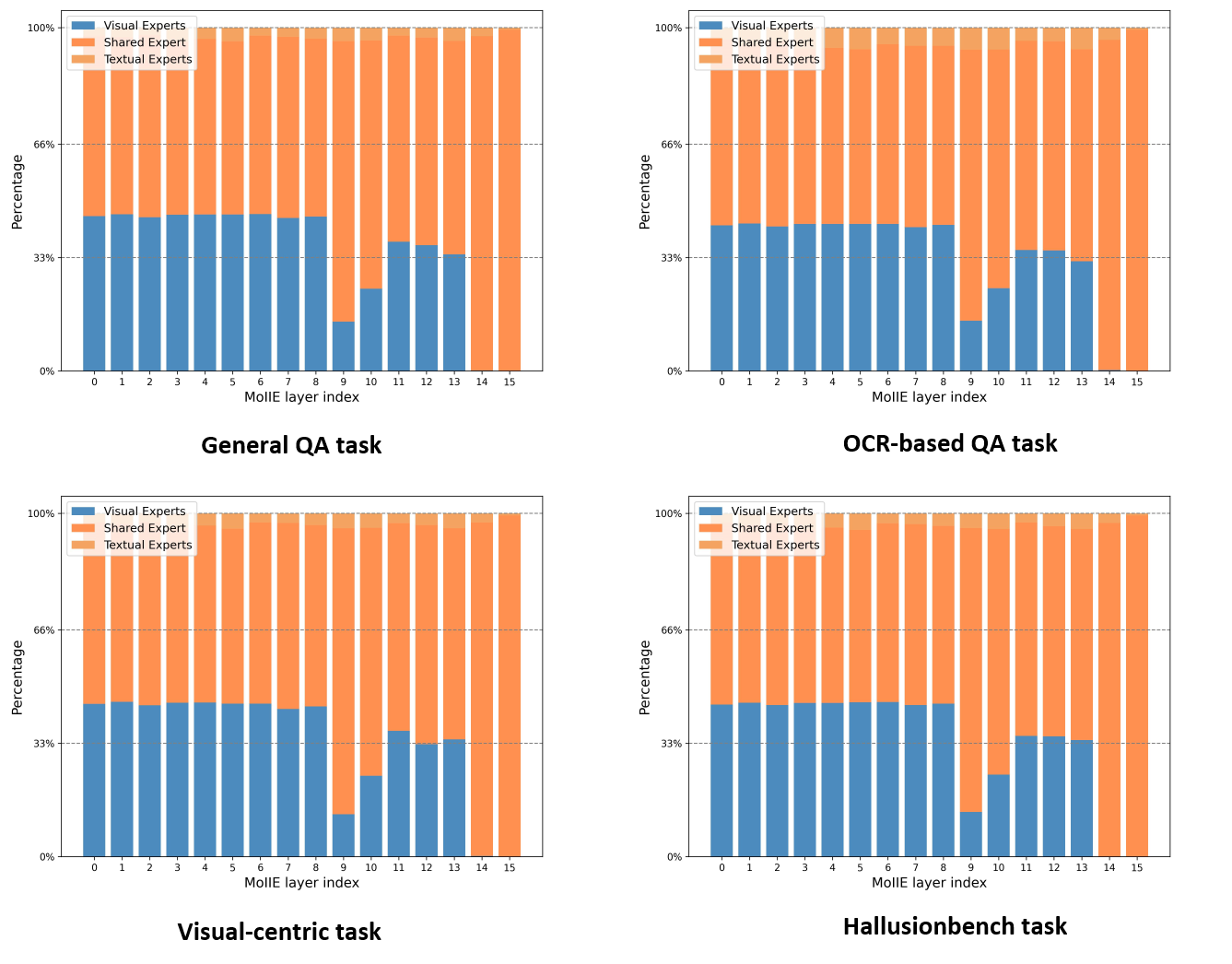}
    \caption{Visualization of the proportion of activated experts across diverse dimensions in multimodal understanding benchmarks.}
    \label{fig:activate}
\end{figure*}

As illustrated in the figure\ref{fig:activate}, we visualize the proportion of activated experts across four representative multimodal understanding benchmarks. For general QA, we use MME~\cite{fu2024mmecomprehensiveevaluationbenchmark}; for OCR-based QA, TextVQA~\cite{singh2019vqamodelsread}; for visual-centric reasoning, we use MMVP~\cite{tong2024eyeswideshutexploring}; and for hallucination, we use HallusionBench~\cite{guan2024hallusionbenchadvanceddiagnosticsuite}.

We consistently observe that lower layers of the model tend to activate modality-specific visual experts, indicating specialized modeling of image features. In contrast, as the layer depth increases, the activation ratio of modality-specific experts decreases, while the use of modality-shared experts increases, suggesting a shift toward modeling shared semantic information across modalities.

\subsection{Training Details}
Our detailed training settings and hyper-parameters of MoIIE are shown in Table~\ref{table:appendix_train}.

\begin{table}[h!]
\resizebox{0.5\textwidth}{!}{
\begin{tabular}{|l|cc|}
\toprule
\textbf{Configuration} & \textbf{Stage 1} & \textbf{Stage 2} \\ \midrule
Experts type & - & FFN \\ \midrule
Experts & - & \begin{tabular}[c]{@{}c@{}}1 visual expert\\ 2 shared experts\\ 1 textual expert\end{tabular} \\ \midrule
Top-K & - & 2 \\ \midrule
Visual encoder & \multicolumn{2}{c|}{siglip-so400m-patch14-384} \\ \midrule
Connection module & \multicolumn{2}{c|}{2 Linear layers with GeLU} \\ \midrule
Image resolution & \multicolumn{2}{c|}{384 x 384} \\ \midrule
Learning rate & \makecell{1e-3\\ \{connection module\}} & \begin{tabular}[c]{@{}c@{}} \makecell{2e-5\\ \{LLM,connection module\}} \\ \makecell{2e-6\\ \{visual encoder\}}\end{tabular} \\ \midrule
LR schedule & \multicolumn{2}{c|}{Cosine decay} \\ \midrule
Weight decay & \multicolumn{2}{c|}{0} \\ \midrule
Optimizer & \multicolumn{2}{c|}{AdamW} \\ \midrule
Warmup ratio & \multicolumn{2}{c|}{0.03} \\ \midrule
Epoch & \multicolumn{2}{c|}{1} \\ \midrule
Global batch size & 256 & 128 \\ \midrule
Deepspeed & Zero2 & Zero3 \\ \midrule
Max token length & \multicolumn{2}{c|}{4096} \\ \midrule

\end{tabular}}
\caption{\textbf{Detailed training hyperparameters of MoIIE.}}
\label{table:appendix_train}
\end{table}

\subsection{Expert Analysis}
\label{sec:Experts Analysis}

\begin{table}[h!]
\centering
\resizebox{.5\textwidth}{!}{
\begin{tabular}{c|ccc}
\toprule
\multicolumn{1}{c|}{} & \multicolumn{1}{c}{MMBench} & \multicolumn{1}{c}{TextVQA} & \multicolumn{1}{c}{MMMU\_val} \\ \midrule
Textual experts       & 72.9                        & 63.1                        & \textbf{41.5}                 \\
Visual experts        & 73.6                        & \textbf{64}                 & 41                            \\
Shared experts        & \textbf{74.1}               & 63.1                        & 41.3                          \\ \midrule
\end{tabular}}
\caption{\textbf{Expert analysis between modality-specific experts and modalitt-shared experts.}}
\label{table:experts_ana}
\end{table}

To evaluate the effectiveness of modality-specific and modality-shared experts, we extracted checkpoints from the shared experts in MoIIE, as well as the modality-specific visual and textual experts from the Modality Expert model. These were tested on the MMbench~\cite{liu2024mmbenchmultimodalmodelallaround} and TextVQA~\cite{singh2019vqamodelsread} and MMMU validation subset~\cite{yue2024mmmumassivemultidisciplinemultimodal} which for General, OCR and konwledge based abilities of model.

As shown in Table~\ref{table:experts_ana} We find that modality-shared experts outperform modality-specific experts on general multi-modal benchmarks like MMbench~\cite{liu2024mmbenchmultimodalmodelallaround}, which require cross-modal reasoning. In contrast, visual experts perform better on OCR-based QA like TextVQA~\cite{singh2019vqamodelsread}, which emphasizes image details and modality-specific information whether texual experts are capable of knowledge-based QA like MMMU validation subset~\cite{yue2024mmmumassivemultidisciplinemultimodal}, which requires the original textual knowledge of model. This further validates the effectiveness of our MoIIE architecture, which demonstrates that successful performance across a wide range of multi-modal tasks requires modeling both cross-modal associations and modality-specific features.

\begin{table*}[!th]
\centering
\caption{\textbf{Comparison of MoIIE with other ablated architectural variants across comprehensive multi-modal benchmarks, including results with different visual encoders and the use of advanced training data such as LLaVA-OV~\cite{li2024llavaonevisioneasyvisualtask}.} “Data” indicates the amount of visual instruction data. SEED-I, HalluB and MMMU$_\text{v}$ are abbreviations for SEED-Image, HallusionBench, and the MMMU validation subset, respectively. Bold numbers represent the best performance in each column.}
\setlength\tabcolsep{2pt}
\resizebox{1\textwidth}{!}{
\begin{tabular}{ccccccccccccccccc}
\toprule
\multicolumn{1}{c|}{} & \multicolumn{1}{c|}{} & \multicolumn{1}{c|}{} &
MMBench & GQA & VQAv2 & MMVet & \multicolumn{1}{c|}{SEED-I} &
POPE & \multicolumn{1}{c|}{HalluB} &
TextVQA & DocVQA & \multicolumn{1}{c|}{ChartQA} &
AI2D & MMMU$_\text{v}$ & \multicolumn{1}{c|}{Mathvista} & \\
\multicolumn{1}{c|}{\multirow{-2}{*}{}} &
\multicolumn{1}{c|}{\multirow{-2}{*}{Data}} &
\multicolumn{1}{c|}{\multirow{-2}{*}{Visual Encoder}} &
\multicolumn{5}{c|}{General Multi-modal QA} &
\multicolumn{2}{c|}{Hallucination} &
\multicolumn{3}{c|}{OCR-based QA} &
\multicolumn{3}{c|}{Knowledge-based QA} &
\multirow{-2}{*}{AVG} \\
\rowcolor[HTML]{DEE0E3}
\multicolumn{17}{c}{\cellcolor[HTML]{DEE0E3}With Advanced Instruction Data} \\ 
\midrule
\multicolumn{1}{c|}{Dense} 
  & \multicolumn{1}{c|}{2.7M} 
  & \multicolumn{1}{c|}{CLIP-L-336} 
  & 74.5 & 63.0 & 81.3 & 40.9 
  & \multicolumn{1}{c|}{69.9} & 86.5 & \multicolumn{1}{c|}{31.2} 
  & 64.1 & 47.7 & \multicolumn{1}{c|}{57.1} 
  & 73.2 & 39.6 & \multicolumn{1}{c|}{30.6} 
  & 58.4 \\
\multicolumn{1}{c|}{Vanilla MoE} 
  & \multicolumn{1}{c|}{2.7M} 
  & \multicolumn{1}{c|}{CLIP-L-336} 
  & 74.6 & 63.1 & 81.1 & 41.0 
  & \multicolumn{1}{c|}{69.7} & 86.5 & \multicolumn{1}{c|}{30.9} 
  & 64.3 & 47.5 & \multicolumn{1}{c|}{57.0} 
  & 74.0 & 40.5 & \multicolumn{1}{c|}{31.0} 
  & 58.5 \\
\multicolumn{1}{c|}{Modality MoE} 
  & \multicolumn{1}{c|}{2.7M} 
  & \multicolumn{1}{c|}{CLIP-L-336} 
  & 74.0 & 63.0 & 81.5 & 39.7 
  & \multicolumn{1}{c|}{69.5} & 86.6 & \multicolumn{1}{c|}{\textbf{32.3}} 
  & 64.4 & 48.1 & \multicolumn{1}{c|}{57.7} 
  & 73.7 & 40.1 & \multicolumn{1}{c|}{31.3} 
  & 58.6 \\
\rowcolor[HTML]{E1EAFF}
\multicolumn{1}{c|}{MoIIE}            
  & \multicolumn{1}{c|}{2.7M} 
  & \multicolumn{1}{c|}{CLIP-L-336} 
  & \textbf{75.5} & \textbf{63.6} & \textbf{82.0} & \textbf{41.9}     
  & \multicolumn{1}{c|}{\textbf{70.7}} & \textbf{87.1} & \multicolumn{1}{c|}{31.7} 
  & \textbf{65.1} & \textbf{48.2} & \multicolumn{1}{c|}{\textbf{58.0}} 
  & \textbf{75.1} & \textbf{41.8} & \multicolumn{1}{c|}{\textbf{32.5}} 
  & \textbf{59.5} \\ 
\midrule
\rowcolor[HTML]{DEE0E3}
\multicolumn{17}{c}{\cellcolor[HTML]{DEE0E3}With Different Visual Encoder} \\ 
\midrule
\multicolumn{1}{c|}{Dense} 
  & \multicolumn{1}{c|}{4M} 
  & \multicolumn{1}{c|}{SigLIP-So400m-384} 
  & 76.2 & 64.6 & 82.5 & 42.3 
  & \multicolumn{1}{c|}{71.4} & 88.1 & \multicolumn{1}{c|}{32.4} 
  & 66.1 & 49.4 & \multicolumn{1}{c|}{58.8} 
  & 75.0 & 41.3 & \multicolumn{1}{c|}{32.1} 
  & 60.0 \\
\multicolumn{1}{c|}{Vanilla MoE} 
  & \multicolumn{1}{c|}{4M} 
  & \multicolumn{1}{c|}{SigLIP-So400m-384} 
  & 76.4 & 64.9 & 82.9 & 42.9 
  & \multicolumn{1}{c|}{71.6} & 88.9 & \multicolumn{1}{c|}{33.8} 
  & 66.2 & 50.2 & \multicolumn{1}{c|}{59.7} 
  & 75.8 & 43.0 & \multicolumn{1}{c|}{33.3} 
  & 60.7 \\
\multicolumn{1}{c|}{Modality MoE} 
  & \multicolumn{1}{c|}{4M} 
  & \multicolumn{1}{c|}{SigLIP-So400m-384} 
  & 75.1 & 64.7 & 82.7 & 41.9 
  & \multicolumn{1}{c|}{71.3} & 88.0 & \multicolumn{1}{c|}{35.3} 
  & 67.4 & 50.8 & \multicolumn{1}{c|}{59.0} 
  & 75.3 & 41.8 & \multicolumn{1}{c|}{32.5} 
  & 60.4 \\
\rowcolor[HTML]{E1EAFF}
\multicolumn{1}{c|}{MoIIE}            
  & \multicolumn{1}{c|}{4M}   
  & \multicolumn{1}{c|}{SigLIP-So400m-384} 
  & \textbf{77.0} & \textbf{65.2} & \textbf{83.4} & \textbf{44.3}    
  & \multicolumn{1}{c|}{\textbf{72.4}} & \textbf{89.7} & \multicolumn{1}{c|}{\textbf{36.0}} 
  & \textbf{69.9} & \textbf{51.0} & \multicolumn{1}{c|}{\textbf{59.9}} 
  & \textbf{77.4} & \textbf{43.5} & \multicolumn{1}{c|}{\textbf{35.3}} 
  & \textbf{61.9} \\ 
\bottomrule
\end{tabular}}
\label{table:extend}
\end{table*}

\subsection{More Experimental Results}
\label{sec:More Experimental Results}
we conduct extended ablation experiments using a different vision encoder(CLIP-L-336) as shown in Table~\ref{table:extend}, replacing the vision encoder does not affect our main conclusions. Under identical training conditions,the proposed MoIIE architecture consistently outperforms all other ablation variants across most of the 13 benchmarks and shows a clear advantage in terms of average performance. 

We also adopt a larger dataset that matches the scale of modern open multimodal benchmarks. Specifically, we followed LLaVA-OV~\cite{li2024llavaonevisioneasyvisualtask}, which provides the high overall data quality among these open datasets a total training corpus of approximately 8M samples, 4M for insturct tuing stage, and used it as our expanded training corpus.

As shown in Table~\ref{table:extend}, using a higher-quality and larger-scale dataset but still identical training conditions across all variants, our proposed MoIIE architecture continues to outperform all other variants across all the 13 multimodal benchmarks. Moreover, MoIIE achieves a more pronounced improvement in average performance (+1.2\%) relative to the other variants. with the trainig data more advanced, the performance gap between MoIIE and other architectures widening.

These results demonstrate that MoIIE not only maintains its advantages under stronger training condition, but also scales more effectively in multimodal training, confirming its robustness and generality.

\end{document}